\pdfoutput=1

\documentclass[11pt]{article}

\usepackage{graphicx}
\usepackage{adjustbox}
\graphicspath{ {./figures/} }
\usepackage{float}
\usepackage{arydshln}
\usepackage{pgfplots}
\usepackage{graphics}
\DeclareGraphicsExtensions{.pdf,.png}

\usepackage{naacl2022}

\usepackage{times}
\usepackage{latexsym}

\usepackage{booktabs}
\usepackage{multirow}

\usepackage[T1]{fontenc}

\usepackage[utf8]{inputenc}

\usepackage{microtype}

\usepackage{amsmath}

\title{An End-to-End Dialogue Summarization System for Sales Calls}

\author{{Abedelkadir Asi$^1$, Song Wang$^2$, Roy Eisenstadt$^1$,Dean Geckt$^1$,} \\\textbf{Yarin Kuper$^1$, Yi Mao$^2$, Royi Ronen$^1$}\\
\textsuperscript{\rm 1}Microsoft Dynamics,\\
\textsuperscript{\rm 2}Microsoft Azure\\
\texttt{\{abeasi,sonwang,reisenstadt,t-deangeckt,} \\
   \texttt{yarinkuper,maoyi,royir\}@microsoft.com}} 

\begin{document}
\maketitle
\begin{abstract}
Summarizing sales calls is a routine task performed manually by salespeople. We present a production system which combines generative models fine-tuned for customer-agent setting, with a human-in-the-loop user experience for an interactive summary curation process. We address challenging aspects of dialogue summarization task in a real-world setting including long input dialogues, content validation, lack of labeled data and quality evaluation. We show how GPT-3 can be leveraged as an offline data labeler to handle training data scarcity and accommodate privacy constraints in an industrial setting. Experiments show significant improvements by our models in tackling the summarization and content validation tasks on public datasets. 
\end{abstract}

\section{Introduction}

An integral part of salespeople daily routine is summarizing sale calls. The summarization process aims to distill salient information from sales dialogues into succinct highlights, which are then leveraged by salespeople for productivity and coaching purposes. Manually curating a call summary is considered as one of the biggest time wasters for B2B sellers~\cite{ZhangTete2020}. It distracts salespeople from nurturing the relationship with their next customer. Recently, this practice has become more demanding due to the emerging landscape of remote selling where virtual meetings become the new norm~\cite{McKinsey2020}.

\begin{figure}[t]
    \centering
    \begin{adjustbox}{max size={\textwidth}{\textheight}}
    \includegraphics[width=0.45\textwidth]{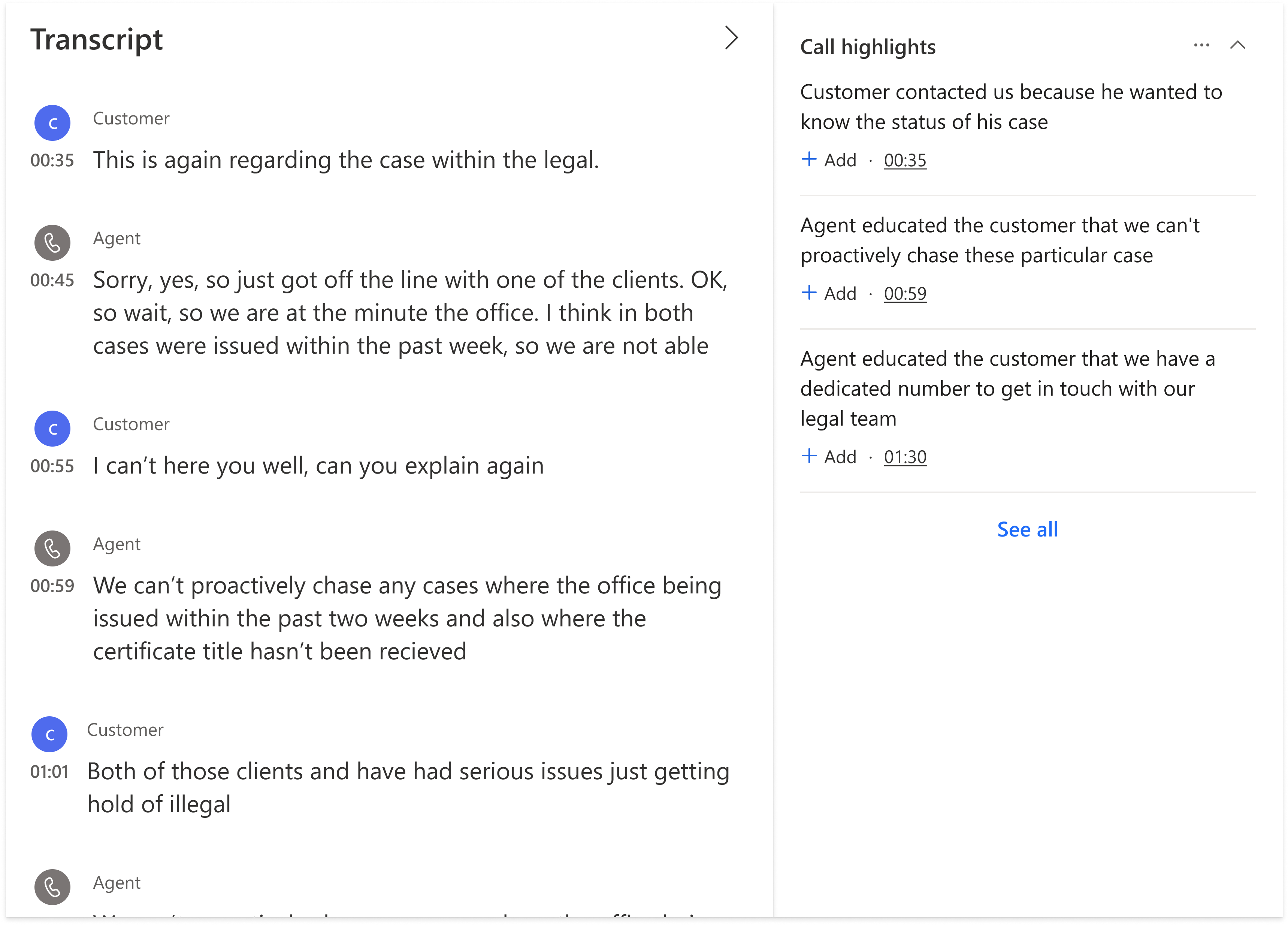}
    \end{adjustbox}
    \caption{A customer-agent call transcript with corresponding summary highlights. Challenges imposed by automatic speech recognition engine can be observed.}
    \label{fig:dialouge_challenges}

\end{figure}

Dialogue summarization induces a variety of unique challenges compared to summarization of documents such as news or scientific papers \citep{zhu2006summarization}. Information density is a key challenge in dialogue text; information is scattered over multiple utterances and participants, leading to frequent coreferences and topic alternations. Spoken dialogues, usually transcribed by speech recognition engines, impose additional challenges such as redundancies and misrecognized words. The length of these dialogues, e.g. $50K$ tokens in a $45$ minutes call, imposes another challenge to state-of-the-art summarization models as it exceeds their input limits~\cite{zhang-etal-2021-exploratory-study}. Figure~\ref{fig:dialouge_challenges} illustrates parts of the challenges imposed by automatically transcribed sales dialogues.

\begin{figure}[ht]
    \centering
    \includegraphics[width=.5\textwidth]{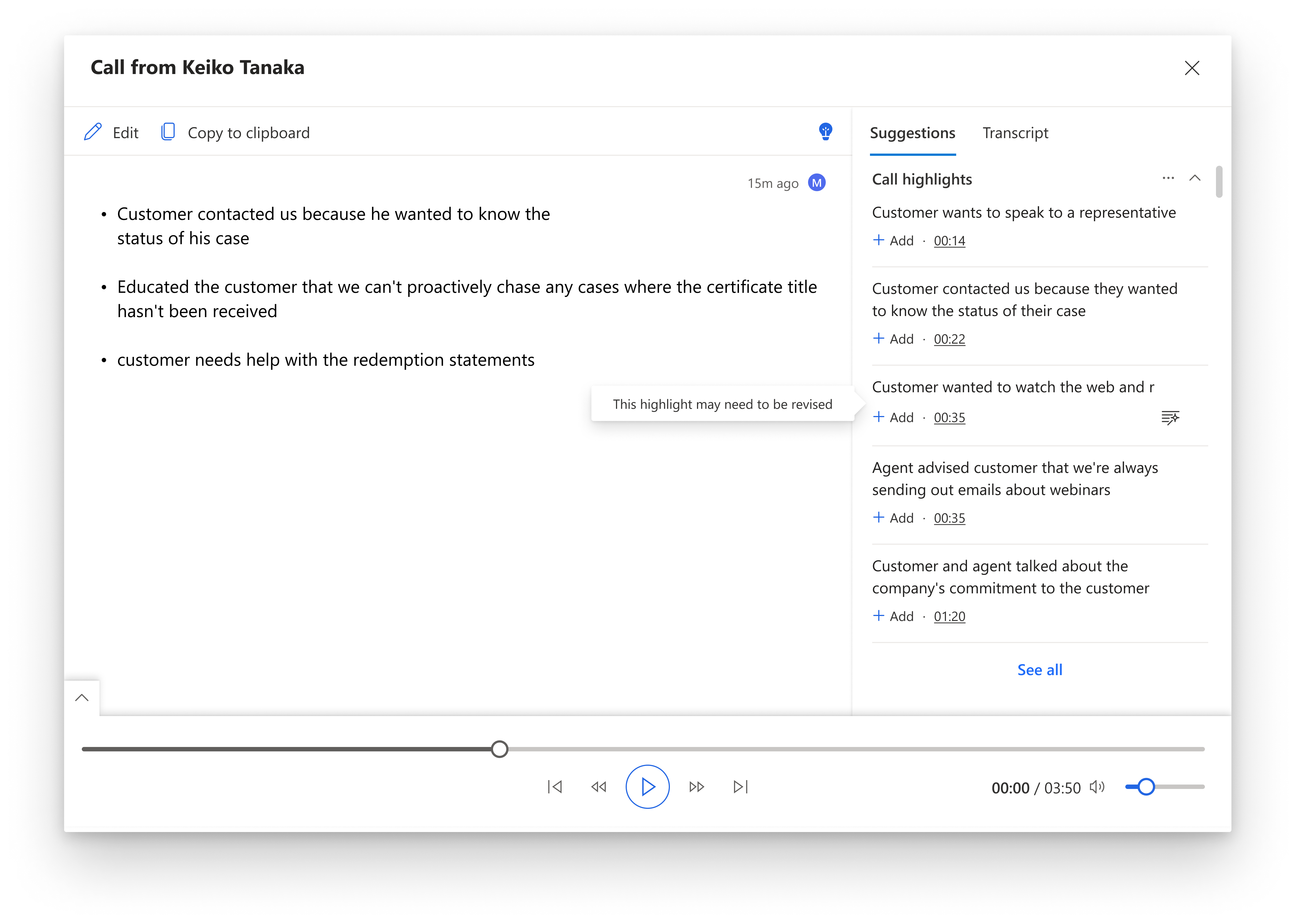}
    \caption{Illustrating human-in-the-loop experience which enables users to interactively handle summarization challenges by adding relevant summary highlights to the editing canvas and modify them, if necessary.}
    \label{fig:system-ui}
\end{figure}

Developing a production system which is both fully automatic and agnostic to the input text genre is an extremely difficult task given the current state-of-the-art technology. To this end, we present a pragmatic solution that enables users to interactively edit machine-generated summary for customer-agent sales calls as appears in Figure~\ref{fig:system-ui}. Our solution summarizes the call into a collection of abstractive highlights to accurately capture the various details of the call. The machine-human interaction is enabled through a designated human-in-the-loop user experience \citep{Ostheimer:2021}. It enables users to modify the generated summaries, yet, the intervention is designed to be minimal so that the overall time consumption of users is significantly reduced. \newline 

Overall, our contributions are listed as follows: 
\begin{enumerate}
    \item \textbf{Dialogue summarization system}. We introduce an innovative production summarization system for summarizing call transcripts with a human-in-the-loop setting. Our system uses an advanced summarization model to generate abstractive summaries for dialogues. Additionally, it employs a novel model for quantifying the coherence of the summaries to compensate for the summarization model limitations.
    \item \textbf{GPT-3 as an offline label generator}. We present a technique for leveraging GPT-3 model to generate pseudo labels without the need to deploy and maintain it in production. This enables us to (i) efficiently generate labels in low-resource setting, (ii) distill GPT-3 knowledge into lighter models, and (iii) accommodate data privacy constraints. 
    \item \textbf{Custom evaluation metric}. We examine the importance of leveraging a comprehensive evaluation metric which takes into account various quality aspects of the generated summary. The metric is utilized to focus our efforts on potential candidate models during the development phase. The suggested metric goes beyond lexical overlap and help us validate that our production model is optimized for generating summaries which are fluent, relevant and factually reliable.     
\end{enumerate}

\section{Related Work}

\paragraph{Document Summarization}
Summarization methods can be categorized into two classes: extractive and abstractive. Early works focused on extractive methods \citep{hovy-lin-1997-automated, marcu-1997-discourse}, followed by rule-based approaches for abstractive summarization \citep{barzilay-elhadad-1997-using, barzilay-etal-1999-information}. Advancements in capabilities of deep neural models led to works such as \citet{rush-etal-2015-neural} where a seq-2-seq attention-based model is used for abstractive summarization. \citet{See2017GetTT} overcomes some of the former work's limitations by introducing a pointer generator network that has the ability to copy words from the source document. A major advancement in the field of deep neural models was the introduction of Transformer architecture \citep{vaswani2017attention}, which is the basis for current state-of-the-art summarization approaches. Recently, several powerful Transformer-based models have been developed and showed remarkable results on various benchmark summarization tasks \citep{LEWIS:2020,zhang2019pegasus, 2020t5} . 

\paragraph{Dialogue Summarization}
The task of dialogue summarization has been witnessing many commonalities as document summarization as well as new techniques for handling unique structures of various dialogue types. Early works in the domain suggested tackling the problem using extractive methods \citep{Murray2005ExtractiveSO, riedhammer2008keyphrase}. 
\citet{shang-etal-2018-unsupervised} used a pure unsupervised graph-based method for keyword extraction and sentence compression. \citet{Goo2018AbstractiveDS} proposed to explicitly model relationships between dialogue acts using attention-based sentence-gated mechanism. \citet{chen-yang-2020-multi} extracted Transformer-based representations for different views of dialogues, conditioned on view representations, to generate summaries using a second Transformer. \citet{zhu-etal-2020-hierarchical} presented a hierarchical Transformer architecture to encompass the structure of dialogues. 

\section{Method}

\begin{figure*}[hbt!]
    \centering
    \includegraphics[width=1.0\textwidth]{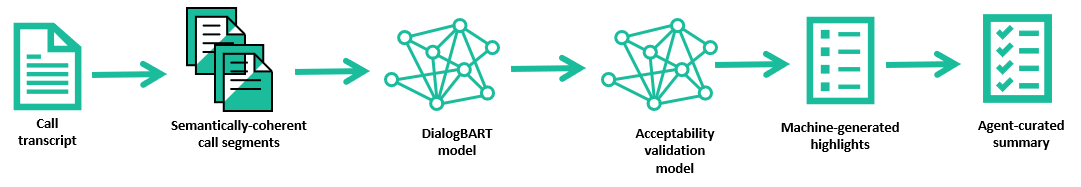}
    \caption{A high-level overview of the system's flow}
    \label{fig:es-system-flow}
\end{figure*}

While most existing methods summarize a call transcript as a single paragraph, our system provides a collection of sentences that summarize the entire dialogue in a chronological order. Given a call transcript, the system utilizes word embeddings to break the transcript into semantically coherent segments \cite{AlemiG15}. Each segment is summarized independently capturing key information such as: customer's issue, agent's solution or the underlying topic of the discussion. Finally, the grammatical coherence of highlights is analyzed using a dedicated model before suggesting them to the user. Figure~\ref{fig:es-system-flow} provides a high-level overview of the system's flow.

Next, we introduce the key components of our dialogue summarization system in details.

\subsection{DialogBART: Dialogue Summarization Model}

Unlike general documents, conversation transcripts have unique structures associated with speakers and turns. In sales calls, participants can either be a customer or an agent and these roles impose a unique language style that can be leveraged by the model. Motivated by this observation, we propose an encoder-decoder model called DialogBART, which adapts the well-known BART \cite{LEWIS:2020} model with additional embedding parameters to model both turns and speakers positions \citep{zhang2019dialogpt,bao-etal-2020-plato}. For speaker embeddings, we introduce designated vectors to represent each speaker which can be easily generalized to multi-participant dialogues. Additionally, we leverage another set of vectors to model turn position embeddings. During inference, the model determines the speaker and turn indices by leveraging a special token that separates the dialogue's turns. 

\begin{figure}[ht!]
    \centering
    \includegraphics[height=2.0in]{"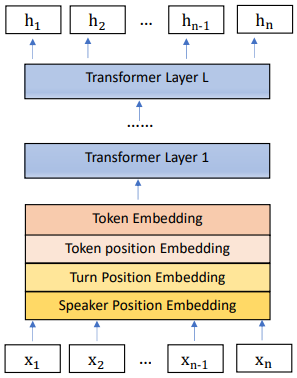"}
    \caption{Input representation of DialogBART's encoder.}
    \label{fig:dialogbart_encoder}
\end{figure}

As shown in Figure \ref{fig:dialogbart_encoder}, DialogBART's input is calculated as the sum of the corresponding token, position, speaker and turn position embeddings. These parameters are randomly initialized, however, the remaining parameters are initialized with weights from a pretrained\footnote{\url{https://huggingface.co/sshleifer/distilbart-xsum-12-3}} BART-like encoder-decoder models~\citep{LEWIS:2020,shleifer2020pretrained}. All these weights are further fine-tuned on dialogue summarization tasks. 

\subsection{Acceptability Validation}

Despite the human-in-the-loop user experience, customers still expect high quality summaries which require minimal modifications by them. We propose a novel model that determines the quality of each summary highlight in terms of coherence, fluency and its acceptability in general.  

\textit{Grammatical acceptability}, a property of natural language text, implies whether a text is accepted or not as part of the language by a native speaker. 

The notion was widely investigated through vast work done in automatic detection of grammatical errors \citep{atwell-1987-detect, Chodorow2000AnUM, bigert2002robust, wagner-etal-2007-comparative} and on acceptability judgment of neural networks \citep{Lau2017GrammaticalityAA, Warstadt2019NeuralNA}. And yet, we are not aware of works that observe the acceptability of neural generated summaries for validation purposes. To determine a highlight's acceptability, we compute the perplexity of each highlight given by a Pretrained Language Model (PLM). This PLM is fine-tuned on summaries from DialogSum dataset \citep{chen-etal-2021-dialogsum} and in-domain proprietary data in a traditional self-supervised manner. Recall that the perplexity of a sequence $W = w_0w_1 ... w_{n}$ is defined as:

\begin{gather}
PP(W;\theta) = \sqrt[n]{\prod_{k=1}^{n}\frac{1}{p_{\theta}(w_k|w_0w_1 ... w_{k-1})}}
\end{gather}

where $\theta$ are the language model specific parameters and $p_{\theta}$ is the probability function corresponding to distribution over vocabulary tokens induced by the same model. 

Based on the perplexity score, the system determines whether a given highlight should be filtered out, presented to the user, or presented with an indication that its revision may be required. Figure~\ref{fig:system-ui} illustrates how the system helps users focus their efforts on modifying borderline acceptable highlights based on the perplexity score. 

\section{GPT-3 as an Offline Labeler}

\begin{figure}[hbt!]
    \centering
    \includegraphics[width=0.5\textwidth]{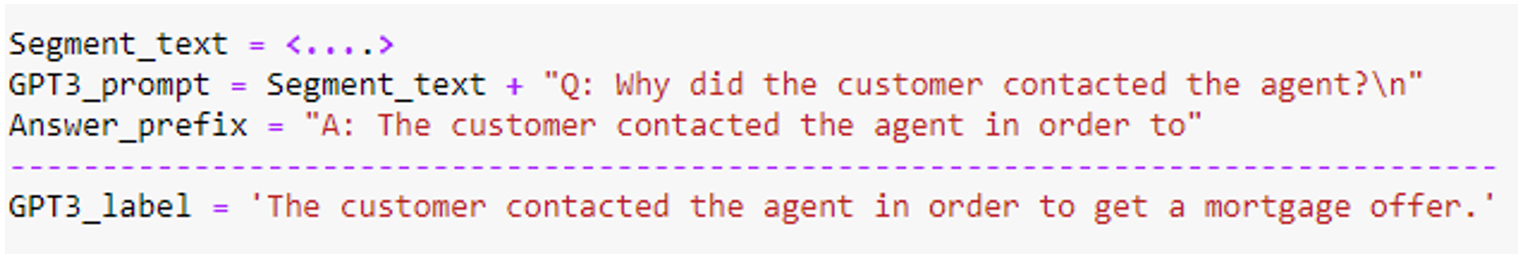}
    \caption{Utilizing GPT-3 model to generate task-oriented summaries in an offline manner.}
    \label{fig:gpt3-prompt}
\end{figure}

Training a dialogue summarization model requires a large amount of labeled examples. Manually annotating data for abstractive summarization is a time-consuming and labor-intensive process, let alone the data privacy constraints. In this work, we provide a method to automatically pseudo label examples to overcome these challenges. We leverage GPT-3 model \cite{NEURIPS2020_1457c0d6} to generate pseudo labels in a zero-shot setting for each call segment. GPT-3 is a large auto-regressive language model with 175 billion parameters that achieves promising results on various NLP tasks, including question answering. We treat the problem of label generation as a question answering task. For each segment, we concatenate a question-based prompt with the segment's content while expecting the GPT-3 model to provide the answer as appears in Figure~\ref{fig:gpt3-prompt}. These answers are used as pseudo labels for the corresponding segments. This formulation provides the flexibility of defining multiple questions per segment to summarize the segment from different perspectives. Finally, these pseudo labels, combined with proprietary human labeled data, are used to fine-tune the DialogBART model on conversational text.

\section{Custom Evaluation Metric}
Common evaluation metrics for text summarization task, i.e. ROUGE and METEOR, have salient limitations as both metrics track lexical overlap between the summary and the original text. This kind of assessment falls short when the summary content perfectly aligns with a reference text but does not necessarily contain any lexical overlap, e.g. abstractive summaries. 

In an industrial setting, one needs to consider various quality perspectives to guarantee that the summary's quality does not introduce productivity blockers for users or negatively affect business decisions. We introduce a custom evaluation metric, \textit{SumSim}, that relies on lexical overlap as well as other quality aspects to ensure that summaries are fluent, coherent and factually reliable. SumSim aims to cover the following quality perspectives:

\begin{itemize}
    \item \textbf{Coverage} - how many units from the reference text are covered by the summary ($S_r$)
    \item \textbf{Relevance} - measures semantic consistency between the summary and the reference text ($S_b$)
    \item \textbf{Informativeness} - how well it captures pre-defined keywords which are critical to the business ($S_i$)
    \item \textbf{Factuality} - how factual the summary is with respect to the original text ($S_f$)  
\end{itemize}

Our metric uses \textit{ROUGE-L} \cite{lin-2004-rouge}, \textit{BertScore} \cite{Tianyi2019}, exact match of keywords and \textit{FactCC} \cite{socher2019} to capture the above quality aspects, respectively. The quality of a given summary is calculated as follows:

\begin{equation}
S_{0} = \alpha \cdot S_i + \frac{1-\alpha}{2} \cdot (S_r + S_b) 
\end{equation}

\begin{equation}
SumSim = \beta \cdot S_f + (1-\beta) \cdot S_0
\end{equation}

where $\alpha$ and $\beta$ are determined empirically based on the business scenario sensitivity. 

\section{Experimental Results}

In this section we evaluate the performance of our proposed models on various datasets: DialogSum \cite{Yulong2021}, SAMSum \cite{SAMSum2019}, CoLA \cite{Warstadt2019NeuralNA} and a proprietary data from the sales domain. We also show the potential of $SumSim$ metric compared to traditional evaluation metrics on the text summarization task. We use Huggingface Transformers \citep{wolf-etal-2020-transformers} as a training framework in all of our experiments.

\subsection{DialogBART}

In the following experiments we show the performance of DialogBART model in summarizing dialogues by examining two factors: (i) speaker/turn position embedding parameters, and (ii) data augmentation by GPT3-labeled data. For comparison purposes, we leverage two baseline models, \textit{BART-large} and \textit{distilBART}, which achieved state-of-the-art results on the summarization task~\cite{LEWIS:2020, shleifer2020pretrained}. All models, including the baseline models, were initially fine-tuned on XSum dataset~\cite{xsum-emnlp}.

First, we examine the contribution of DialogBART's position embeddings on DialogSum and SAMSum datasets. All models were fine-tuned using the relevant training sets and evaluated on the test sets of the corresponding datasets. Table~\ref{tab:dialogbart_embedding} shows that the suggested speakers/turns positions embeddings provide better results when compared to the baseline models. 

\begin{table}[!ht]
\centering
\begin{tabular}{*4l}
\toprule
{Model} & \textbf{R1} & \textbf{R2} & \textbf{RL}\\
\textbf{} & \multicolumn{3}{c}{\textbf{DialogSum}} \\

\midrule
distilBART & 35.93 & 11.71 & 28.86\\
\hspace{5pt} + embeddings & \textbf{46.97} & \textbf{21.34} & \textbf{39.45} \\
\hdashline
BART-large& 46.48 & 20.89 & 38.12\\
\hspace{5pt} + embeddings & \textbf{46.68} & \textbf{21.46} &\textbf{38.32}\\
\midrule
\textbf{} & \multicolumn{3}{c}{\textbf{SAMSum}} \\
\midrule
distilBART  & 41.93 & 19.17 & 34.05\\
\hspace{5pt} + embeddings & \textbf{50.21} & \textbf{25.89} & \textbf{41.99} \\
\hdashline
BART-large &52.45 &	28.08	&43.84\\
\hspace{5pt} + embeddings & \textbf{52.91} & \textbf{28.39} & \textbf{43.90} \\

\bottomrule
\end{tabular}
\caption{Effectiveness of DialogBART's speaker and turn embedding parameters using ROUGE metrics.}
\label{tab:dialogbart_embedding}
\end{table}

Second, we examine the implications of fine-tuning DialogBART model using different data types: human-labeled ($20K$ samples) and GPT3-labeled ($21K$ samples) data \footnote{The anonymized data was collected and used based on a data sharing agreement with customers from different business domains. The human-labeled data is composed of anonymized agent's notes which were captured as part of the daily routine of the agents and not in a crowdsourcing setting.}.

 We evaluated the models on the test subset of: (i) DialogSum (500 samples), (ii) SAMSum (819 samples), and (iii) proprietary data (100 samples). The evaluation on the public datasets was conducted without fine-tuning the models on the corresponding training sets. Table~\ref{tab:dialogbart_benchmarking} shows that DialogBART model outperforms the baseline models on public datasets even in out-of-domain setting. Additionally, results show that DialogBART fine-tuned on a mixture of human and pseudo labels outperforms its counterparts which were fine-tuned on either fully human labels or fully pseudo labels. We note that fine-tuning DialogBART on pseudo labels yielded higher ROUGE scores compared to human labels. This could be explained by human tendency to generate variable summaries which induces disagreements between human annotators \citep{Clark2021AllT}. While a model fine-tuned on pseudo labels is less variable in its generations, a model fine-tuned using human data produces text that is, in turn, more variable and leads to less lexical overlap with test references as measured by ROUGE metrics\footnote{\url{https://github.com/google-research/google-research/tree/master/rouge}} . 

\begin{table*}
\centering
\begin{tabular}{llllllllll}
\toprule
\textbf{Model} & \multicolumn{3}{c}{\textbf{DialogSum}} &  \multicolumn{3}{c}{\textbf{SAMSum}} & \multicolumn{3}{c}{\textbf{Proprietary}} \\
{} & \textbf{R1} & \textbf{R2} & \textbf{RL} &  \textbf{R1} & \textbf{R2} & \textbf{RL} & \textbf{R1} & \textbf{R2} & \textbf{RL} \\
\midrule
distilBART & 17.1 & 3.6 & 13.5 & 20.3 & 4.1 & 15.5 & 16.3 & 1.1 & 13.1 \\
BART-large & 17.7 & 3.9 & 13.7 & 24.9 & 5.5 & 18.9 & 16.9 & 1.5 & 13.3 \\
\hdashline
DialogBART\\
\hspace{5pt} + human & 21.7 & 4.5 & 19.1 & 18.9 & 3.0 & 16.8 & 20.2 & 5.3 & 19.2 \\
\hspace{5pt} + pseudo & 28.0 & 5.9 & 22.2 & 26.0 & 5.1 & 20.6 & 28.5 & 10.5 & 24.8 \\
\hspace{5pt} + human \& pseudo & \textbf{33.5} & \textbf{9.0} & \textbf{24.5} & \textbf{30.4} & \textbf{7.5} & \textbf{22.4} & \textbf{33.1} & \textbf{13.0} & \textbf{26.3}\\
\bottomrule
\end{tabular}
\caption{Results of fine-tuning DialogBART model on human labels, pseudo labels and mixture of them.}
\label{tab:dialogbart_benchmarking}
\end{table*}

\subsection{Acceptability Validation}

We examine mutiple candidate PLMs with language model objective for this task. Initially we fine-tune the candidate PLM on summaries from DialogSum dataset and later on positive examples from the train subset of our internal acceptability benchmark consisting of in-domain summaries ($100$ samples). As candidate PLMs, we experiment with GPT-2 \citep{radford2019language}, DistilGPT-2 \citet{sanh2019distilbert} and a RoBERTa encoder \citep{Liu2019RoBERTaAR} with a language model head, \textit{RoBERTa-LM}. Table \ref{tab:cola_acceptability} shows comparison results between the examined models and leaderboard competitors on the development set of CoLA as well as on the test subset of an internal benchmark. 

\begin{table}[ht!]
    \centering
    \resizebox{\columnwidth}{!}{%
    \begin{tabular}{lcc}
    \toprule
        \textbf{Model} & \textbf{CoLA\textsubscript{dev}} & \textbf{Internal}\\
        \midrule
        TinyBERT \citep{jiao-etal-2020-tinybert} & 54 & -\\
        Synthesizer \citep{Tay2021SynthesizerRS} & 53.3 & -\\
        DeBERTa \citep{he2021deberta} & \textbf{69.5} & 54.5\\
        \midrule
        DistilGPT-2 & 61.7 & 63.6\\
        GPT-2 & 62.5 & 60.6\\
        RoBERTa-LM & 64.2 & \textbf{66.7}\\
        \bottomrule
    \end{tabular}%
    }
    \caption{Accuracy of acceptability validation models.}
    \label{tab:cola_acceptability}
\end{table}

We observe that all of the models trained using our method, in the bottom half of the table, which were not trained on CoLA data yield competitive results compared to models explicitly fine-tuned for the task, top half of Table \ref{tab:cola_acceptability}. We also found that the RoBERTa-LM model achieves highest results on the internal set. Additionally, we fine-tuned DeBERTa, the strongest CoLA competitor, in a classification setting on the internal benchmark. We observe that results achieved by our models are significantly better. We hypothesize, this phenomenon is due to the fact that valid in-domain highlights, as generated by DialogBART, share a unique structure and can be viewed as forming a specific language which properties are better captured by a language model rather than a classifier.

\subsection{Custom Evaluation Metric}
\begin{figure}[!hbt]
    \centering
    \includegraphics[width=.45\textwidth]{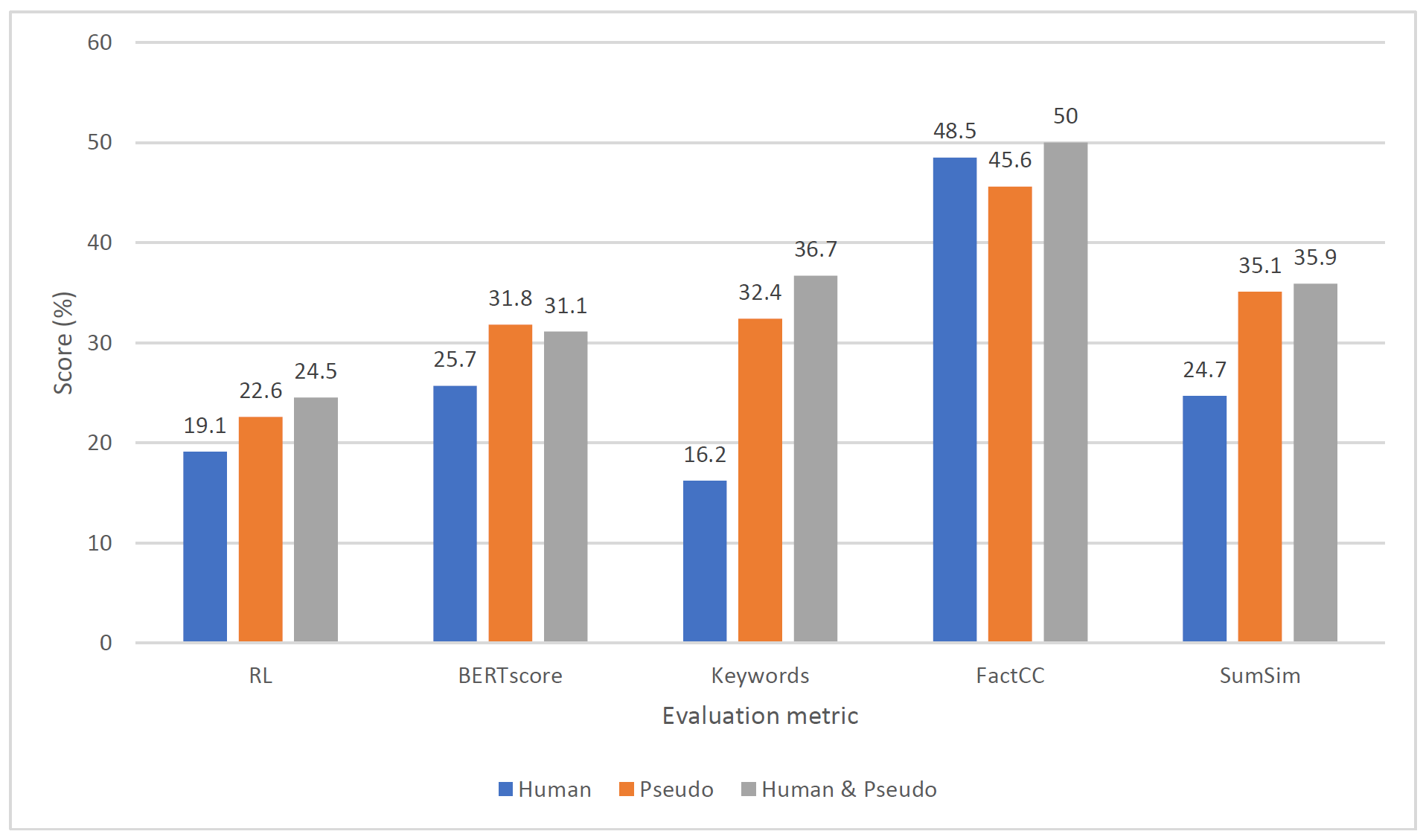}
    \caption{The capacity and limitation of various quality metrics. The bars' colors represent three different models which were fine-tuned on human labels, pseudo labels and a mixture of them, respectively.}
    \label{fig:metric-breakdown}
\end{figure}

We leverage the DialougSum test set to show the potential of the $SumSim$ metric. Table~\ref{tab:dialogbart_benchmarking} shows that DialogBART fine-tuned on pseudo labels, $M_{pseudo}$, outperforms its counterpart that was fine-tuned on human labels, $M_{human}$. However, Figure~\ref{fig:metric-breakdown} shows contradicting insights when comparing the performance of these two models by different quality aspects, i.e., lexical overlap (ROUGE) and factual reliability (factCC).      

This observation emphasizes the need for non-standard quality measures for evaluating the performance of abstractive summarization models. This need is critical for customer-facing enterprise products where business decisions can be affected by the generated summary. Figure~\ref{fig:metric-breakdown} shows the strengths and weaknesses of different quality metrics in evaluating three DialogBART variants. The $M_{pseudo}$ model outperforms the $M_{human}$ in all quality dimensions except of factuality. This observation is consistent with recent studies which report that large size language models are less truthful than their smaller peers~\cite{Lin2021TruthfulQAMH}.        

\section{Conclusions}
In this paper, we present an end-to-end system for abstractive summarization of agent-customer calls. We employ a two-stage strategy to summarize long call transcripts by (i) splitting the dialogue into semantically coherent segments, and (ii) generating summaries using our DialogBART summarization model. We demonstrate how a pragmatic solution that combines a content selection model with a human-in-the-loop user experience can help compensate on generative models' limitations. We show how GPT-3 model can be leveraged as an offline data labeler to train lighter models and accommodate data privacy constraints. Experiments show that the introduced embedding parameters combined with fine-tuning on in-domain data significantly improve the quality of the generated summaries with respect to publicly available BART-based summarization models. We emphasize the need for non-standard evaluation metrics and show how common metrics fall short when evaluation of abstractive summaries is considered.          

\bibliography{es_naacl2022}
\bibliographystyle{acl_natbib}

\end{document}